\pdfoutput=1

\documentclass[11pt]{article}

\usepackage{naacl2021}

\usepackage{times}
\usepackage{latexsym}

\usepackage[T1]{fontenc}

\usepackage[utf8]{inputenc}

\usepackage{microtype}

\usepackage{url}

\usepackage{booktabs}

\usepackage{xspace,mfirstuc,tabulary}
\usepackage{graphicx}
\usepackage{subcaption}

\newcommand{\textblue}[1]{\textcolor{black}{#1}}
\newcommand{\textred}[1]{\textcolor{red}{#1}}
\newcommand{\secref}[1]{\S\ref{sec:#1}}

\usepackage{multicol}
\usepackage{multirow}
\usepackage{kky}
\usepackage{url}
\usepackage{algorithm}
\usepackage[noend]{algpseudocode}
\usepackage{cleveref}
\crefformat{section}{\S#2#1#3} 
\crefformat{subsection}{\S#2#1#3}
\crefformat{subsubsection}{\S#2#1#3}

\usepackage{stackengine,scalerel}

\newcommand\overstar[1]{\ThisStyle{\ensurestackMath{%
  \setbox0=\hbox{$\SavedStyle#1$}%
  \stackengine{0pt}{\copy0}{\kern.2\ht0\smash{\SavedStyle*}}{O}{c}{F}{T}{S}}}}

\newcommand{\occ}[1]{\text{occ}(#1)}

\title{Phrase-level Active Learning for Neural Machine Translation}

\author{
 Junjie Hu, Graham Neubig \\
 Language Technologies Institute, Carnegie Mellon University \\
  {\tt \{junjieh,gneubig\}@cs.cmu.edu}\\
}

\date{}

\begin{document}
\maketitle
\begin{abstract}
Neural machine translation (NMT) is sensitive to domain shift. In this paper, we address this problem in an active learning setting where we can spend a given budget on translating in-domain data, and gradually fine-tune a pre-trained out-of-domain NMT model on the newly translated data. Existing active learning methods for NMT usually select sentences based on uncertainty scores, but these methods require costly translation of full sentences even when only one or two key phrases within the sentence are informative. 
To address this limitation, we re-examine previous work from the phrase-based machine translation (PBMT) era that selected not full sentences, but rather individual phrases.
However, while incorporating these phrases into PBMT systems was relatively simple, it is less trivial for NMT systems, which need to be trained on full sequences to capture larger structural properties of sentences unique to the new domain.
To overcome these hurdles, we propose to select \emph{both} full sentences and individual phrases from unlabelled data in the new domain for routing to human translators.
In a German-English translation task, our active learning approach achieves consistent improvements over uncertainty-based sentence selection methods, 
improving up to 1.2 BLEU score over strong active learning baselines.%
\footnote{All code/data will be released upon acceptance.}
\end{abstract}

\section{Introduction}

\begin{figure}
    \centering
    \includegraphics[width=0.93\columnwidth]{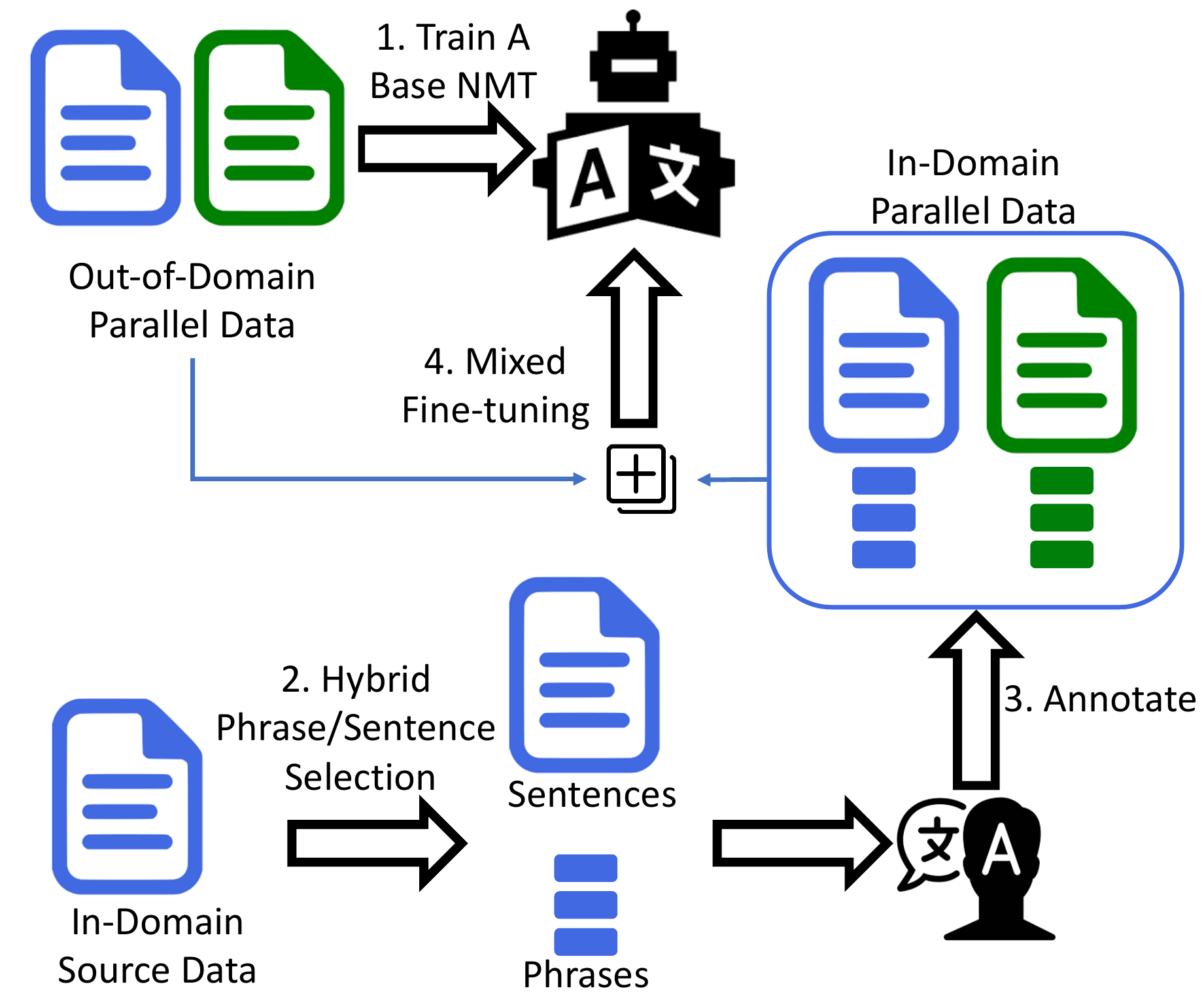}
    \vspace{-0.1cm}
    \caption{Overview of the active learning process}
    \label{fig:my_label}
    \vspace{-0.4cm}
\end{figure}

Machine translation (MT) models are very sensitive to domain shift~\cite{koehn-knowles-2017-six,chu-wang-2018-survey}, and one typical way to address this problem is adding in-domain data to the MT training process~\cite{Luong-Manning:iwslt15,chu-etal-2017-empirical}.
However, this data may not be available \textit{a priori}, and hiring professional translators with knowledge of specific domains (such as medicine or law) is usually costly.

As a result, active learning approaches~\cite{gangadharaiah-etal-2009-active,haffari-etal-2009-active,bloodgood-callison-burch-2010-bucking} have been widely adopted to reduce the annotation cost by translating a smaller representative subset of the in-domain data, with the hope that models trained on this translated subset approximate those trained on a much larger labeled set. 
In general, active learning (AL) approaches iterate between two steps: \emph{data selection/annotation}, and \emph{model update}. With regards to data selection for machine translation, most existing works \cite{haffari-etal-2009-active,peris-casacuberta-2018-active,zeng-etal-2019-empirical} focus on selecting \emph{sentences} that are most useful for training either phrase-based machine translation (PBMT) or neural machine translation (NMT) models.

However, even the most informative sentences inevitably involve segments that the MT system can already translate well, and asking the translator to also translate these segments is not cost-effective. There have been a few works used in conjunction with older PBMT models that ameliorate this problem through phrase-based selection techniques~\cite{bloodgood-callison-burch-2010-bucking,daume-iii-jagarlamudi-2011-domain,miura-etal-2016-selecting}, which select only \emph{individual phrases}, maximizing information gain.
However, while these translated phrases can be easily integrated into PBMT by adding them to the existing phrase table, incorporating them into NMT models is less simple because NMT has no concept of a ``phrase table'' and must be trained on full sentences similar to those that must be translated.

In this paper, we propose a method for incorporating phrase-based active learning into NMT.
Specifically, we first describe sentence-based and phrase-based selection strategies, then propose a hybrid strategy that combines both methods.
We also describe several ways to incorporate this translated data into the training of NMT systems.
We conduct experiments on German-English translation by adapting NMT models trained on WMT parallel data to the medicine and IT domains. 
Experimental results show that the hybrid selection strategy obtains more stable translation performance than either phrase-based or sentence-based selection strategy. 


\section{Problem Definition}
In the setting of active learning for domain adaptation, we are given an out-of-domain labelled corpus $(x, y) \in \Lcal$ and an in-domain unlabelled corpus $x \in \Ucal$.  We define a phrase as a contiguous sequence of words up to some length limit $N$, and denote a set of possible phrases in a sentence $x$ by $\cup_{n\in[1,N]} n\text{-gram}(x)$, where we set $N=4$ in all experiments below. To obtain translations of unlabelled data, we assume access to professional translators $\Ocal(\cdot)$ who can translate source-side sentences $\Scal$ and/or phrases $\Pcal$ selected from $\Ucal$, i.e., $\Ocal(x)~\forall x\in\Scal \subset\Ucal$, and $\Ocal(p)~\forall p\in \Pcal \subset \Pcal_{\Ucal}=\cup_{x\in \Ucal} \cup_{n\in[1,N]} n\text{-gram}(x)$. We assume that translating sentences or phrases requires cost $c(\cdot)$, and annotation must be performed within a fixed budget $B=\sum_{x\in \Scal}c(x) + \sum_{p\in \Pcal} c(p)$. This active learning procedure consists of two main steps: selection/translation (\cref{sec:selection}) and fine-tuning (\cref{sec:fine-tune}).

\section{Active Selection Strategies} \label{sec:selection}

\subsection{Sentence Selection Strategies}

Existing sentence-based active learning methods usually define a sentence-level scoring function $\phi(x,\cdot)$, and select sentences with the top scores. Following \citet{zeng-etal-2019-empirical}, we categorize these methods into two classes: data-driven and model-driven methods. Data-driven methods only rely on the unlabeled data $\Ucal$ and the labeled data $\Lcal$, i.e., $\phi(x, \Ucal, \Lcal)$, and usually score sentences based on the trade-off between the density and diversity of the selected sentences. In contrast, model-driven approaches usually estimate the prediction uncertainty of a source sentence given the current MT model $\theta$, i.e., $\phi(x, \theta, \Ucal, \Lcal)$, and select sentences with high uncertainty for training the model. Before getting to our proposed phrase-based strategies in \secref{phrase-select} we highlight several existing sentence selection strategies.

\paragraph{Random Sampling:} One easy strategy is randomly sampling sentences from the unlabeled data $\Ucal$ for annotation. Although it is simple, this method is an unbiased approximation of the data distribution in $\Ucal$. Therefore, this method remains a strong baseline in the active learning literature~\cite{gangadharaiah-etal-2009-active,miura-etal-2016-selecting,zeng-etal-2019-empirical} if the annotation budget is sufficiently large.

\paragraph{Cosine Similarity between Sentence Embeddings (CSSE):} \citet{zhang2018active} propose to measure the distance between sentence embeddings. This method takes each unlabeled sentence, estimates its distance in embedding space from the labeled sentences in the out-of-domain corpus, and iteratively selects sentences that are more distant from sentences in the labeled data. In our instantiation of this method, we leverage the pre-trained mBERT model \cite{devlin-etal-2019-bert} to extract sentence representation $\eb_x$ of a particular sentence $x$,%
\footnote{
We average the word representations from the 7th layer of the mBERT model as the sentence embedding, because the middle-layer representations have proven effective in cross-lingual retrieval tasks \cite{pires-etal-2019-multilingual,hu2020}.
}
and measure a ratio-based distance \cite{artetxe-schwenk-2019-margin} which is the ratio between the cosine similarity of $(\eb_x, \eb_{x'})$ and the average cosine similarity with their $k$ nearest neighbors:
\vspace{-0.3cm}
\begin{align} \label{eq:ratio-dist}
    &\text{ratio}(\eb_x, \eb_{x'})  \\ \nonumber
    &~= \frac{ \cos{(\eb_x, \eb_{x'})} }{ \sum\limits_{z\in \text{NN}_k(x)} \frac{ \cos{(\eb_x, \eb_z)} }{2k} +  \sum\limits_{z\in \text{NN}_k(x')} \frac{ \cos{(\eb_{x'}, \eb_z)} }{2k}  },
\end{align}
\vspace{-0.5cm}

\noindent where $k$ is the number of nearest neighbors. 

We then compute the margin-based distance between each in-domain sentence and its nearest out-of-domain neighbor within a randomly sampled subset of labeled sentences $\Lcal'$:
\begin{align}
    \label{eq:csse}
    &\phi(x, \cdot) = \text{dist}(x, \Lcal') = \min_{x'\in \Lcal'} \text{ratio}(\eb_x, \eb_{x'}).  
\end{align}

We approximate the distance between $x$ and out-of-domain corpus $\Lcal$ using a subset $\Lcal'$ for efficiency purposes, because the out-of-domain $\Lcal$ is usually large. Next we use the minimal distance $\text{dist}(x, \Lcal')$ as our scoring function $\phi(x,\cdot)$, and select the unlabeled sentence with the largest distance from (sub-sampled) sentences in the out-of-domain corpus.

\paragraph{Round Trip Translation Likelihood (RTTL):} One model-driven method is based on a method referred to as ``round trip translation'' \cite{haffari-etal-2009-active,zeng-etal-2019-empirical}. The labeled data $\Lcal$ is used to train two MT models $\theta_\text{src-tgt}, \theta_\text{tgt-src}$ that translate between the source and target languages in two directions. Each unlabeled source sentence $x\in\Ucal$ is first translated to $\hat{y}$ in the target language by $\theta_\text{src-tgt}$, and then $\hat{y}$ is translated to $\hat{x}$ by $\theta_\text{tgt-src}$. This method assumes that if this round-trip translation process fails to recover some of the content on the source side then this is an indication that the sentence may be difficult for the current model and is a good candidate for human annotation. \citet{haffari-etal-2009-active} use a scoring function that computes the similarity between the original sentence $x$ and $\hat{x}$ using the sentence-level BLEU score \cite{chen-cherry-2014-systematic}, while \citet{zeng-etal-2019-empirical} estimate the likelihood of the original source sentence $x$ given $\hat{y}$ by the reverse MT model $\theta_\text{tgt-src}$. 
\vspace{-0.2cm}
\begin{align}
    &\hat{y} \approx \argmax_y  P_{\theta_\text{src-tgt}}(y|x)  \\
    &\phi(x, \cdot) = \log P_{\theta_\text{tgt-src}}(x| \hat{y})
\end{align}

\subsection{Phrase Selection Strategies} \label{sec:phrase-select}

A few existing phrase-based active learning methods~\cite{bloodgood-callison-burch-2010-bucking,miura-etal-2016-selecting} have been proposed to improve PBMT systems. These methods first determine the possible set of phrases in a sentence, select phrases to be translated according to a scoring metric, and incorporate these in the training of the PBMT system. In the following paragraphs, we introduce two phrase-based selection strategies, and discuss how to integrate this data into NMT in \secref{fine-tune}.

\paragraph{$n$-gram Frequency (NGF)}~\cite{bloodgood-callison-burch-2010-bucking}:
The most straightforward phrase selection strategy is to select the most frequent phrases in the unlabelled data that \emph{do not} appear in the already labeled data. First we extract two sets of possible $n$-grams $(n\leq 4)$ from sentences in $\Ucal$ and $\Lcal$, which are defined as $\Pcal_{\Ucal} = \cup_{x\in \Ucal} \cup_{n\in[1,N]} n\text{-gram}(x)$, and $\Pcal_{\Lcal} = \cup_{(x,y)\in \Lcal} \cup_{n\in[1,N]} n\text{-gram}(x)$. We then select the most frequent in-domain phrases from $\Pcal_U$ by 
\vspace{-0.3cm}
\begin{align} \label{eq:hng}
    p = \argmax_{p\in \Pcal_{\Ucal}, p \notin \Pcal_{\Lcal}} \occ{p, \Ucal} ,
\end{align}
\vspace{-0.4cm}

\noindent where $\occ{p}$ counts the occurrences of $p$ in $\Ucal$.

\paragraph{Semi-Maximal Phrases (NGF-SMP):}  The two phrase sets $\Pcal_{\Ucal}, \Pcal_{\Lcal}$ extracted by the $n$-gram Frequency method contain many substrings that also occur in some longer strings. For example, $p$ = ``eines der'' always co-occurs with
the longer $p'$ = ``eines der besten'' in the WMT14 German-English dataset. To identify the longer strings, \citet{miura-etal-2016-selecting} proposed the following semi-order relation, which defines the relation between a phrase $p'$ and its sub-string $p$ satisfying the condition that $p'$ occurs at least half the time of $p$ in the corpus $\Ucal$.
\vspace{-0.4cm}
\begin{align} \nonumber
    p \overstar{\preceq} p' \Leftrightarrow \exists \alpha,\beta: \alpha p \beta = p' \wedge \frac{\occ{p, \Ucal}}{2} < \occ{p', \Ucal}
\end{align} 
\vspace{-0.4cm}

A phrase $p$ is called a semi-maximal phrase if there does not exist a phrase $p'$ in $\Ucal$ such that $p\overstar{\preceq} p'$. Therefore, a compact subset of phrases $\Pcal'_{\Ucal}$ can be constructed by containing only semi-maximal phrases in the phrase set $\Pcal_{\Ucal}$ in $\Ucal$:
\begin{align} \label{eq:phrase-subset}
    \Pcal'_{\Ucal} = \{p| \nexists p'\in \Pcal_{\Ucal}, p\overstar{\preceq} p' \wedge p\in \Pcal_{\Ucal}\} .
\end{align}
\vspace{-0.25cm}

By using semi-maximal phrases in $\Pcal'_{\Ucal}$ rather than all phrases in $\Pcal_{\Ucal}$, we remove a large number of phrases that are included in a longer phrase more than half the time, and reduce the redundancy of the selected phrases. Next we can select phrases similarly using Eq.~(\ref{eq:hng}) by replacing the original phrase set $\Pcal_{\Ucal}$ with the sub-set $\Pcal'_{\Ucal}$.

\subsection{Hybrid Selection Strategy}
Phrase-based selection has its benefits, such as efficient annotation of core vocabulary from the target domain.
However, at the same time it lacks the ability to identify larger sentence structure that may nonetheless be unique to the target domain.
Modeling this structure is particularly important for NMT (in constrast to PBMT), as NMT directly learns both lexical and syntactic transformations within the same model.

Because of this, we propose a simple yet novel hybrid selection strategy that leverages the benefits of both sentence-based and phrase-based selection strategies. Specifically, we allocate our budget in a way to annotate sentences with $B_s$ words from our set of sentences and $B_p$ words from our set of phrases.
Depending on the specific sentence-based and phrase-based selection strategies chosen in the hybrid selection strategy, it is non-trivial to determine which selection strategy improves the in-domain translation performance more than the other one before actual finetuning.  Therefore, in our implementation, we assume that we have no prior knowledge about which selection strategies will be most effective, and simply evenly distribute the annotation budget into the sentence-based and phrase-based strategies. We leave more sophisticated allocation strategies as future work, and we discuss some potential avenues briefly in~\secref{conclusion}.

\section{Training with Sentences and Phrases} \label{sec:fine-tune}


After data selection, we fine-tune the base NMT model on the newly translated data. This is essentially an extreme form of domain adaptation where we adapt a base NMT model trained on out-of-domain data to a new domain. 
Specifically, we adapt a strategy of \emph{mixed fine-tuning} \citep{Luong-Manning:iwslt15}, which continues training a pre-trained out-of-domain model on both in-domain data and a certain amount of out-of-domain data to prevent overfitting to relatively small in-domain data.
Compared to the standard domain adaptation setting where we have only a small number of in-domain sentences, our phrase-level active learning setting has the additional difficulty of having to use short translations of individual phrases.
In the following we describe both methods to choose which data to use in mixed fine-tuning, and how to incorporate phrasal translations. 

\subsection{Data Mixing}


For data mixing, we sample a subset $\Lcal_r$ of data directly from the labeled set $\Lcal'$, and concatenate $\Lcal_r$ with the newly annotated sentences $\Lcal_s$ and phrases $\Lcal_p$ for mixed fine-tuning (Line~\ref{alg:line:fine-tune} in Algorithm~\ref{alg:active}). Specifically, we define a distribution function $\psi$ over $\Lcal'$, and either sample by $(x,y)\sim \psi$ or greedily take the most likely data by $(x,y)=\argmax_{(x,y)\in\Lcal'} \psi(x,y)$ iteratively for $M$ times to obtain the subset $\Lcal_r$ of $M$ parallel data.

\paragraph{Random Sampling:}
The most simple way to select out-of-domain data is to randomly sample sentences from the out-of-domain corpus $\Lcal'$, i.e., $(x,y)\sim \text{Uniform}(\Lcal')$. Although it is simple, this has been popularly used in the literature of domain adaption for NMT~\cite{chu-wang-2018-survey}.  

\paragraph{Retrieve Similar Sentences:}
Recently \citet{aharoni-goldberg-2020-unsupervised} showed that pre-trained language models implicitly learn sentence embeddings that cluster by domains, and proposed a data selection method that has proven more effective than methods based on the likelihood of an in-domain language model~\cite{moore-lewis-2010-intelligent}. Since our base NMT model is pre-trained on out-of-domain corpus, we need to adapt the model to the domain of the unlabeled data. Instead of random sampling, we adopt the selection method in \citet{aharoni-goldberg-2020-unsupervised} to retrieve parallel sentences from $\Lcal'$ that are close to the in-domain sentences in $\Ucal$. To do so, we leverage the contextualized sentence representations, and measure the distance of a source sentence in $\Lcal'$ w.r.t. the unlabeled corpus $\Ucal$ by $\text{ratio}(x, \Ucal),~\forall x\in \Lcal'$. Next, we iteratively retrieve labeled data from $\Lcal'$ that have the smallest distance scores to their nearest neighbors, i.e., $(x,y) = \argmax_{(x,y)\in \Lcal'} \text{ratio}(x, \Ucal)$. 

\subsection{Incorporating Phrasal Translations}


In addition to obtaining real parallel data from $\Lcal'$ for mixed fine-tuning, we create synthetic parallel data $(\hat{x}, \hat{y})$ by incorporating phrasal translations into existing context from $\Lcal'$. Specifically, for an unlabeled sentence $x\in \Ucal$ containing a newly annotated phrase $p_x$, we retrieve the similar sentence pair $(x^*, y^*)$ from $\Lcal'$ by 
\begin{align} \label{eq:swith-argmax}
    & (x^*, y^*)=\argmax_{(x', y')\in\Lcal'}\text{ratio}(\eb_x, \eb_{x'}) 
\end{align}

We then alter $(x^*,y^*)$ with the newly annotated phrase pair $(p_x,p_y)$ to create synthetic sentence pair $(\hat{x}, \hat{y})$. Similar to data mixing, we concatenate the set of synthetic data with the annotated sentences $\Lcal_s$ and phrases $\Lcal_p$ for mixed fine-tuning. 

\paragraph{Switch Phrases:} Inspired by existing data augmentation methods~\cite{fadaee-etal-2017-data}, we examine a data augmentation method that switches out phrases in the out-of-domain sentence pairs in $\Lcal'$ by the newly annotated phrase pairs from $\Ucal$. First, we define the following operation $\text{Switch}(x, p, i)$ that returns a new sentence by substituting the phrase at the $i$-th position in $x^*$ by $p_x$.

\vspace{-0.25cm}
\begin{align}
    \text{Switch}(x^*, p_x, i) = [x^*_{<i} ; p_x ; x^*_{\geq i+|p|}] 
\end{align}
\vspace{-0.25cm}

Next, we enumerate all possible positions in $x^*$ for switching phrases, and 
then apply the in-domain language model trained on $\Ucal$ to select the most probably synthetic sentence by

\vspace{-0.8cm}
\begin{align} \nonumber
    & \hat{x} = \argmax_{\substack{x'=\text{Switch}(x^*, p_x, i) \\ \forall 0 \leq i < |x^*|-|p|,~~ p_x \in \cup_{n\in[1,N]}n\text{-gram}(x)}} P_\text{LM}(x')
\end{align}
\vspace{-0.5cm}

\noindent where $p_x$ is a phrase in the unlabeled sentence $x$.

To synthesize the corresponding $\hat{y}$ from the retrieved target sentence $y^*$, we apply a word alignment model trained on $\Lcal$ to find the index $j$ for the translation of the  replaced phrase $x^*_{i:i+|p_x|}$ in $y^*$, and substitute the phrase at the $j$-th position in $y^*$ by $p_y$ to obtain $\hat{y}=\text{Switch}(y^*, p_y, j)$. 

\paragraph{Contextualized Phrases:} The other idea is to augment the context of a newly annotated phrase pair $(p_x, p_y)$, since a phrase $p_x$ lacks larger sentence structure. Specifically, we define the contextualized operation that augments a phrase $p_x$ in $x$ by appending it to the retrieved sentence $x^*$.
\vspace{-0.2cm}
\begin{align} \label{eq:context_cat}
    & \text{Contextualize}(x^*, p_x) = [x^*, p_x] 
\end{align}
\vspace{-0.25cm}

We then enumerate all annotated phrases in $x$, and apply an in-domain language model to find the most probable annotated phrase pair $(p_x, p_y)$ that synthesizes $\hat{x}$. The corresponding $\hat{y}$ can be obtained by $\text{Contextualize}(y^*, p_y)$.
\vspace{-0.2cm}
\begin{align} \label{eq:context_lm}
    \hat{x} = \argmax_{\substack{x'=[x^*, p_x] \\ \forall p_x\in \cup_{n\in[1,N]}n\text{-gram}(x)}} P_\text{LM}(x')
\end{align}
\vspace{-0.4cm}

\section{Experiments}
\subsection{Experimental Setting}
We use the WMT14 German-English data as the out-of-domain labeled data for training our base NMT model, and take the source sentences of two parallel corpora in the medicine and IT domains~\cite{koehn-knowles-2017-six} as the unlabeled data. More details can be found in Appendix~\ref{sec:exp_details}.

As our NMT model, we use a 6-layer 512-unit Transformer~\cite{NIPS2017_7181} implemented in \texttt{Fairseq},\footnote{\url{https://github.com/pytorch/fairseq}} and use a subword vocabulary of 5,000 for both languages constructed by Byte Pair Encoding \cite{sennrich-etal-2016-neural}.
We train the base model with Adam for 10 epochs with 4K warmup steps and a peak learning rate of 1e-3, and decay the learning rate based on the inverse square root of the number of update steps \cite{NIPS2017_7181}. 

For active learning, we set our annotation budgets by number of words translated (following the prevailing translation market practice to charge for jobs by the word), and investigate the budgets from 2.5K words up to 40K words.%
\footnote{At current market rates, this would cost from 491 to 7,092 USD for German-English translation by professional translators at \url{https://translated.com/}.}
After data selection (\secref{selection}), we obtain a set $\Lcal_r$ of $M$ parallel sentences (\secref{fine-tune}), and set the size $M=|\Lcal_p|$ where $\Lcal_p$ is selected by NGF-SMP. We then fix $\Lcal_r$ for mixed fine-tuning in all experiments, and continue fine-tuning the base model on a mixture of the newly-translated data and $\Lcal_r$ for 5 more epochs. 

\begin{figure*}
    \centering
    \begin{subfigure}{0.42\textwidth}
        \includegraphics[width=\linewidth]{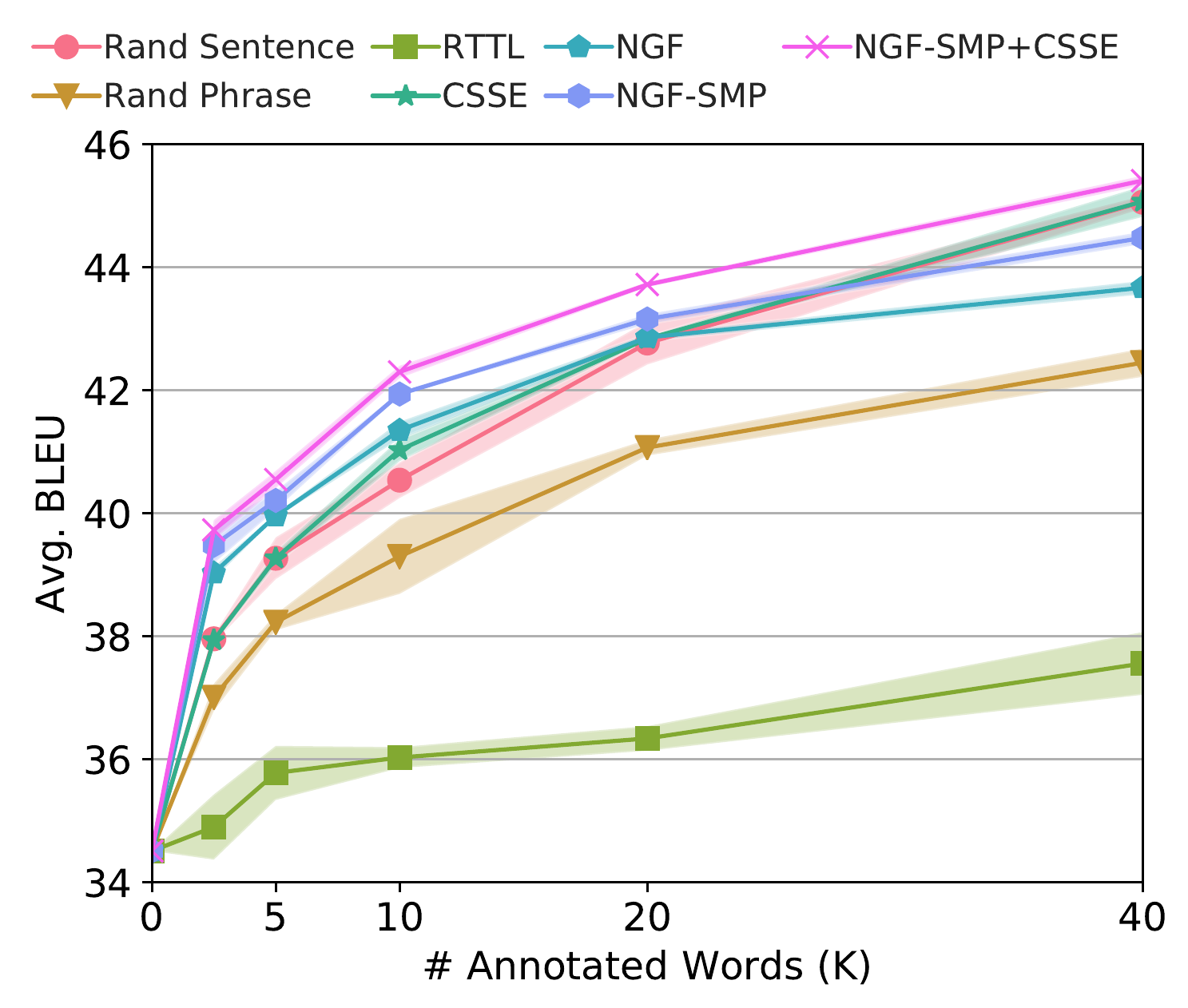}
          \caption{WMT14$\rightarrow$Medicine}
          \label{fig:wmt2emea}
      \end{subfigure}
    \begin{subfigure}{0.42\textwidth}
        \includegraphics[width=\linewidth]{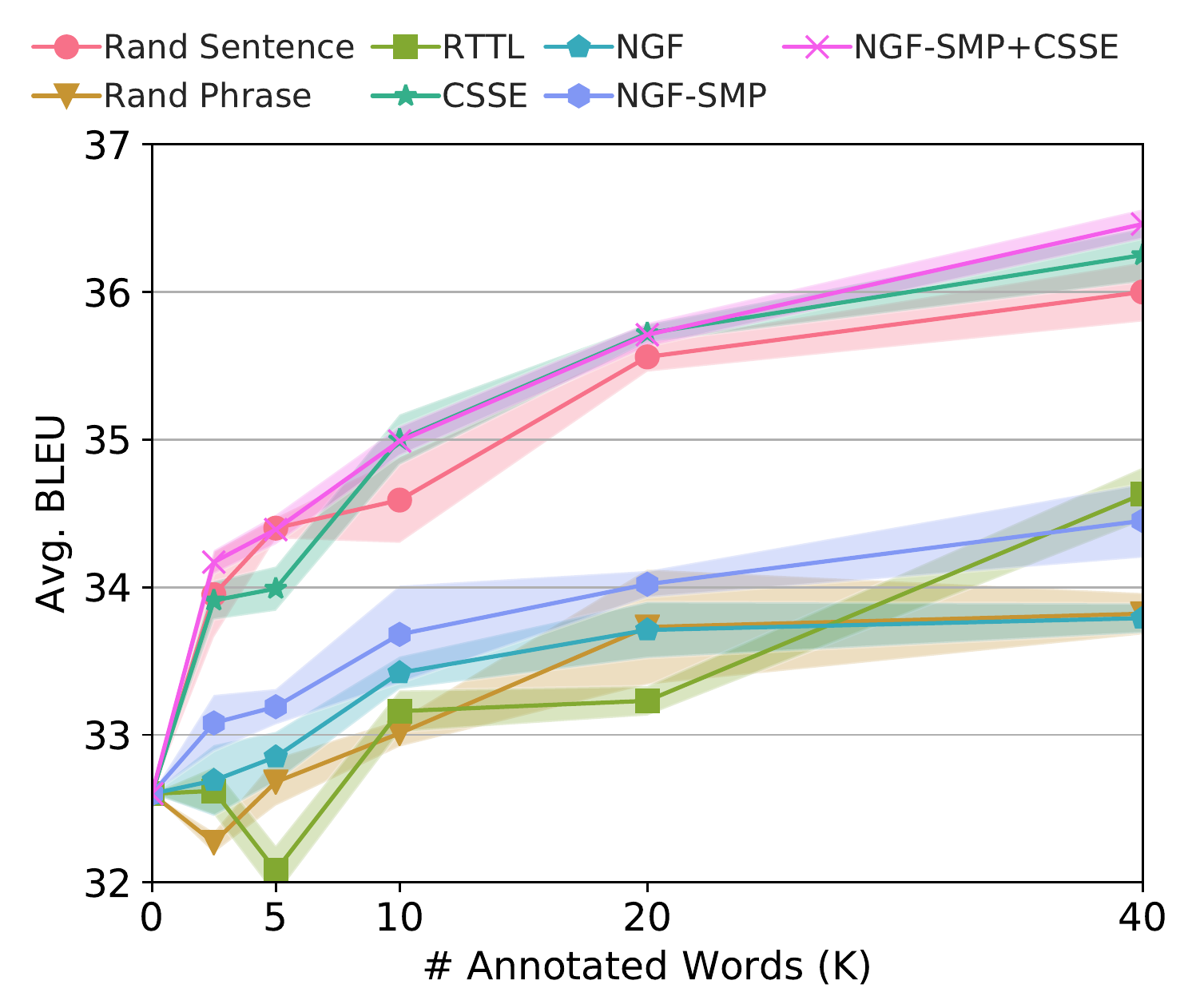}
          \caption{WMT14$\rightarrow$IT}
          \label{fig:wmt2it}
      \end{subfigure}
      \vspace{-0.35cm}
    \caption{Average BLEU score over 3 runs for adapting a base NMT to the Medicine and IT domains.}
    \label{fig:selection}
    \vspace{-0.2cm}
\end{figure*}




\subsection{Word-level Translation Accuracy}
\begin{figure}[h]
    \centering
    \includegraphics[width=0.97\columnwidth]{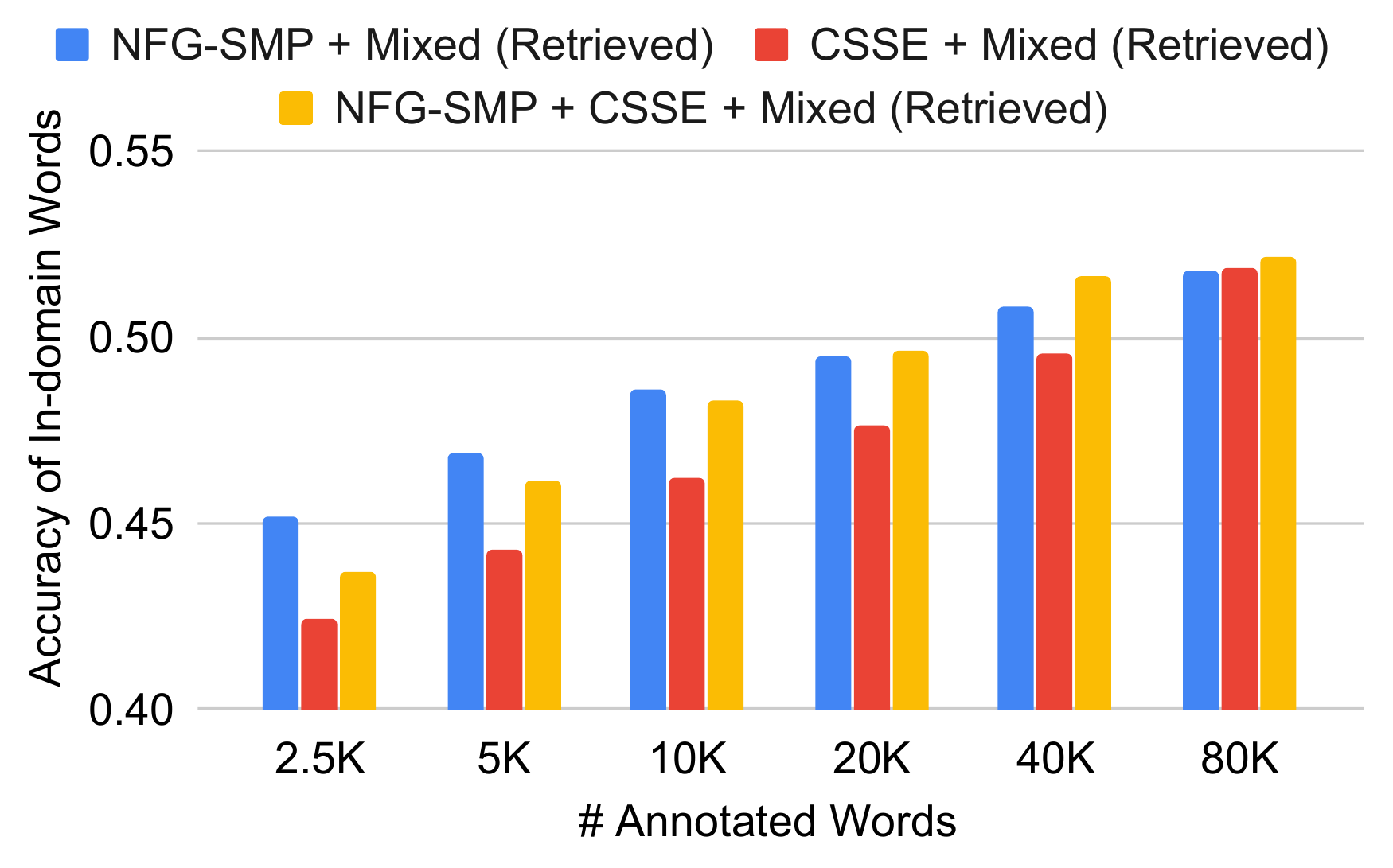}
    \vspace{-0.3cm}
    \caption{Translation accuracy of in-domain words in the test set from the medicine domain}
    \vspace{-0.5cm}
    \label{fig:accuracy}
\end{figure}

Since our selection and mixed fine-tuning methods focus on leveraging phrasal translations for domain adaptation, we perform a fine-grained analysis on the word-level translation accuracy of the NMT systems due to the domain shift.
A source word is defined as an unseen in-domain word when it never appears in the out-of-domain corpus. If phrase selection strategies select more in-domain words, we would expect a higher translation accuracy of such in-domain words by the adapted NMT systems using phrase selection. As a result, we compare the translation accuracy of in-domain words by the NMT models using different selection strategies in Figure~\ref{fig:accuracy}. As shown in the figure, NGF-SMP significantly improves the translation accuracy of the in-domain words with a small annotation budget. In contrast, CSSE falls short of the other compared methods when the annotation budget is less than 80K words. Moreover, we find that the hybrid selection strategy of NGF-SMP and CSSE can combine the merits of both methods, and obtain an even higher accuracy when the budget is greater than 40K annotated words. Qualitatively, the example in Table~\ref{tab:case_study} shows the translations for a source sentence with all words appearing in the medical domain. The NMT model adapted by CSSE translates the first half of the source sentence by picking the correct word ``exercised'', while the NMT model adapted by NGF-SMP generates the correction translation ``somnolence'' in the second half of the output. The NMT model using the hybrid of NGF-SMP and CSSE strategies translates both words correctly (more examples in Appendix~\ref{sec:qualitative_sup}).

\begin{table*}[]
    \resizebox{0.98\textwidth}{!}{%
    \centering
    \begin{tabular}{l|l@{\hskip -0.5cm}r} \toprule 
         & Output & S-BLEU \\ \midrule
        \multirow{2}{*}{Source}  & Jedoch ist Vorsicht geboten, da Berichten zufolge Verwirrung und Somnolenz während der Behandlung & \\ & auftreten können. &  \\
        Reference & However, caution should be exercised as confusion and somnolence have been reported. &  \\
        NGF-SMP & However, caution \textred{is required}, as \textred{there are reports of} confusion and somnolence \textred{during the treatment.} & 15.71 \\
        CCSE & However, caution should be exercised, as confusion and \textred{drowsiness may occur during the treatment.} & 15.62\\
        NGF-SMP+CSSE & However, caution should be exercised as confusion and somnolence \textred{may occur during the treatment.} & 15.71 \\ \bottomrule
    \end{tabular}}
    \vspace{-0.2cm}
    \caption{Translations generated by NMT models using different selection strategies. The last column shows the sentence BLEU score of the translations. Translation errors are highlighted in red.}
    \vspace{-0.4cm}
    \label{tab:case_study}
\end{table*}

\subsection{How Does Each Selection Strategy Help?}

We examine the question of which selection strategy (\secref{selection}) best improves accuracy on in-domain test data.
For mixed fine-tuning, in this section we use the retrieved out-of-domain parallel data for a fair comparison among all active selection strategies. 
Figure~\ref{fig:selection} shows the average BLEU score and the standard deviation of the adapted MT systems to two new domains over 3 independent runs.%
\footnote{To obtain a stable result, we independently run the active learning procedure with different selection strategies 3 times, collect new translation data, and concatenate them with the same data retrieved from out-of-domain labeled data}

Comparing among sentence selection strategies in Figure~\ref{fig:selection}, CSSE performs slightly better than the random sentence selection baseline on adapting the NMT model to the IT domain with smaller standard deviation values, and performs comparably on adapting to the medicine domain. However, we observe that RTTL performs worst, 
and we conjecture that this is due to the usage of the base NMT models that are trained on the out-of-domain parallel data in both directions. The errors accumulated from the round trip translation process lead to an inaccurate estimation of the uncertainty score for a source sentence. Table~\ref{tab:rttl} shows the top 5 sentences selected by RTTL. The selected sentences in the medicine domain are short phrases rather than complete sentences, and those selected in the IT domain contain duplicate phrases such as ``bewerten mit\^{a}''. 
\begin{table}[th!]
\vspace{-0.45cm}
\resizebox{0.48\textwidth}{!}{%
    \centering
    \begin{tabular}{l@{\hskip 2mm}l}\toprule
         \multirow{5}{*}{MED} &Portugal Lundbeck Portugal Lda Quinta da Fonte Edifício D. \\
         & Bronchitis \\
         &Gastrointestinaltrakt : \\
         &Neugebore \\
         &139 B. \\ \midrule
        \multirow{5}{*}{IT}& Eigenschaften des Stichwortes \^{a} \% 1\^{a} \\
&bewerten mit\^{a} Drei Sternen\^{a}\\
&keine Speicherplatzinformation auf\^{a} procfs\^{a} \\
&bewerten mit\^{a} Einem Stern\^{a}\\
&neue und einzelne ausw\~{A} \^{A} hlen\\ \bottomrule
    \end{tabular}}
    \vspace{-0.1cm}
    \caption{Top 5 sentences selected by RTTL}
    \vspace{-0.45cm}
    \label{tab:rttl}
\end{table}

For phrase-based selection methods, NGF-SMP significantly outperforms the random phrase selection strategy. Further, NGF-SMP even outperforms sentence selection methods when the annotation budget is small (less than 20k words) for adaption to the medicine domain.
As we increase the annotation budget to 40K annotated words, sentence selection strategies outperform phrase selection strategies. This indicates that if we keep training NMT systems on shorter phrase pairs when the annotation budget is sufficient, the NMT systems would be limited by lack of longer sentence structures. In Figure~\ref{fig:wmt2it}, we also find that NMT models trained with phrasal translations fall short of those trained with sentence translations when adapting to the IT domain. It is hard to train the NMT systems to translate certain phrases correctly without the sentence context. For example, ``Persönlichen Ordner'' in the IT domain is translated to ``home directory'' rather than ``personal folder'' in the sentence ``jedes Skript dieses Dialogs hat Schreib-Zugriff auf Ihren Persönlichen Ordner ''. 

Finally, the hybrid selection of NGF-SMP and CSSE strategies outperforms the individual selection strategies over every budget in our set of budgets, i.e., 2.5K, 5K, 10K, 20K, 40K annotated words, improving the best phrase selection strategy NGF-SMP by 0.49 average BLEU points, and the best sentence selection strategy CSSE by 1.11 average BLEU points in the medicine domain. Notably, the phrase-based selection strategy especially helps in the scenario where the context is not required to translate domain-specific words, for example, the name of a medicine or a disease in the medicine domain (See the first example in Appendix~\ref{sec:qualitative_sup}). For the adaptation scenario that requires a longer context in some domains such as IT, the hybrid strategy can also significantly outperforms the best phrase-based strategy NGF-SMP by 1.2 average BLEU points, and the best sentence selection strategy CSSE by 0.15 BLEU points. Overall, our hybrid selection strategy is effective to combine the merits of both sentence and phrase selection strategies in the domain adaptation setting.  

\subsection{How Representative Are the Selected Data?}
If the selected data has a significant overlap of segments with the in-domain test data, we would expect a better adaptation performance of the NMT trained on the selected data. Therefore we investigate the $n$-gram overlap between the selected data and the test data when we annotate 5K words from the medicine corpus, and report the average BLEU score of the adapted NMT models trained on the selected data in Table~\ref{tab:ngram}. Interestingly, we find that there exists a high correlation ($\rho\approx$0.8) between the $n$-gram overlap and the average BLEU score, which indicates that the $n$-gram overlap with the test set can be used as a good measure of whether the selected data is useful for improving the NMT model in the new domain. Compared to the random phrase selection, NGF-SMP selects phrases with a high overlap with the test data. We also observe that sentence selection strategies cover fewer phrases in the test data than phrase selection strategies. This also corroborates our assumption that asking translators to annotate phrases that the MT system can already translate well is not cost-effective to improve the in-domain translation performance. 

\begin{table}[]
\centering
\resizebox{0.49\textwidth}{!}{%
\begin{tabular}{lrrrrr} \toprule
Methods         & uni-gram & bi-gram & tri-gram & 4-gram & Avg. BLEU\\ \midrule
OoD Data    & 79.33 & 32.65 & 7.30 & 1.10 & 34.51\\
~+ Random Sentence & 82.81 & 38.45 & 11.62 & 3.73 & 39.27 \\
~+ RTTL    & 80.70 & 35.76 & 9.85 & 3.04 & 35.78\\
~+ CSSE    & 82.74 & 38.83 & 12.01 & 4.05 & 39.27\\ 
~+ Random Phrase   & 82.36 & 35.84 & 7.98 & 1.15 & 38.23\\
~+ NGF & 84.45 & 41.82 & 14.94 & 6.17 & 39.96 \\
~+ NGF-SMP  & 85.80 & 43.13 & 16.15 & 7.11 & 40.21\\
~+ NGF-SMP + CSSE   & 84.48 & 41.89 & 14.98 & 6.48 & 40.55\\
ID Training Data    & 98.58 & 87.30 & 67.61 & 52.11 & 57.59\\ \midrule
Pearson Correlation & 0.90 & 0.83 & 0.80 & 0.78 & / \\\bottomrule
\end{tabular}%
}
\vspace{-0.2cm}
\caption{Percentage of the n-gram in the test sentences that are covered by the selected data with 5K words, the out-of-domain training data and the in-domain training data. The last row shows the Pearson correlation coefficient between $n$-gram overlap and avg. BLEU score.}
\vspace{-0.5cm}
\label{tab:ngram}
\end{table}

\begin{table*}[]
\centering
\resizebox{0.98\textwidth}{!}{%
\begin{tabular}{cccc|cc|ccccc} \toprule
\multicolumn{4}{c|}{Out-of-domain Data}                & \multicolumn{2}{c|}{In-domain Data}                    & \multirow{2}{*}{2.5K}                & \multirow{2}{*}{5K}                  & \multirow{2}{*}{10K}                 & \multirow{2}{*}{20K}                 & \multirow{2}{*}{40K}                 \\
Sampled            & Retrieved & Switched & Contextualized & NGF-SMP                       & CSSE                     &                                      &                                      &                                      &                                      &                                      \\  \midrule
                          &                           & & & \checkmark &                           & \multicolumn{1}{l}{39.39 $\pm$ 0.14} & \multicolumn{1}{l}{39.22 $\pm$ 0.00} & \multicolumn{1}{l}{40.56 $\pm$ 0.02} & \multicolumn{1}{l}{41.19 $\pm$ 0.25} & \multicolumn{1}{l}{44.07 $\pm$ 0.33} \\
                          &                           & & &                          & \checkmark & \multicolumn{1}{l}{37.94 $\pm$ 0.08} & \multicolumn{1}{l}{38.68 $\pm$ 0.54} & \multicolumn{1}{l}{40.62 $\pm$ 0.59} & \multicolumn{1}{l}{42.62 $\pm$ 0.03} & \multicolumn{1}{l}{45.00 $\pm$ 0.11} \\
                          &                           & & & \checkmark & \checkmark & 38.94 $\pm$ 0.02                     & 39.60 $\pm$ 0.09                     & 41.34 $\pm$ 0.12                     & 42.44 $\pm$ 0.15                     & 44.90 $\pm$ 0.06                     \\ \midrule
\checkmark &                           & & & \checkmark & \checkmark & 39.46 $\pm$ 0.14            & 40.51 $\pm$ 0.23                     & 40.62 $\pm$ 0.49                     & 41.82 $\pm$ 0.26                     & 43.78 $\pm$ 0.57                     \\
                          & \checkmark & & & \checkmark & \checkmark & \textbf{39.73} $\pm$ 0.16 & 40.55 $\pm$ 0.14 & \textbf{42.30} $\pm$ 0.10 & \textbf{43.72} $\pm$ 0.04 & \textbf{45.41} $\pm$ 0.08  \\
                          &  & \checkmark & & \checkmark & \checkmark & 38.93 $\pm$ 0.36 &      40.59 $\pm$ 0.17 &      41.82 $\pm$ 0.29 &      42.70 $\pm$ 0.37 &      45.33 $\pm$ 0.04  \\ 
                          & & & \checkmark & \checkmark & \checkmark &  35.36 $\pm$ 0.38 &      37.85 $\pm$ 0.68 &      39.96 $\pm$ 0.35 &      42.83 $\pm$ 0.11 &      44.14 $\pm$ 0.15 \\ 
                          & \checkmark & \checkmark & & \checkmark & \checkmark & 39.61 $\pm$ 0.06 &      \textbf{40.95} $\pm$ 0.06 &      42.19 $\pm$ 0.08 &      43.42 $\pm$ 0.17 &      45.06 $\pm$ 0.19  \\ 
                          & \checkmark & & \checkmark & \checkmark & \checkmark & 37.88 $\pm$ 0.25 &      39.52 $\pm$ 0.32 &      41.17 $\pm$ 0.28 &      42.80 $\pm$ 0.21 &      44.28 $\pm$ 0.13  \\ \bottomrule
\end{tabular}%
}
\vspace{-0.2cm}
\caption{Comparison between mixed fine-tuning methods. Bold indicates highest average BLEU by column.}
\vspace{-0.4cm}
\label{tab:mixed-finetune}
\end{table*}

\subsection{How Do Phrasal Translations Help in Mixed Fine-tuning?}
We further investigate the effect of mixed fine-tuning using the newly annotated in-domain data and sub-sampled out-of-domain data when comparing with fine-tuning only on the newly annotated data. Table~\ref{tab:mixed-finetune} shows the average BLEU score and the standard deviation values over 3 independent runs. Compared to fine-tuning on only annotated data, adding randomly sampled sentence pairs from the out-of-domain data helps when the annotation budget is less than 5K annotated words, but hurts when we increase the budget. In contrast, adding sentences retrieved by the similarity in the sentence embedding space not only outperforms fine-tuning only on annotated data and mixed fine-tuning with randomly sampled sentences, but also achieve smaller standard deviation values. On the other hand, mixed fine-tuning on synthetic data by switching phrases performs slightly worse than the mixed fine-tuning on real retrieved data, but outperform the fine-tuning without any out-of-domain data, especially when the annotation budget is small, e.g., 5K annotated words. Combining synthetic data by switching phrase and real retrieved data for mixed fine-tuning also improve the translation performance over the training only on synthetic data. However, the contextualized method performs worst among all mixed fine-tuning methods, which indicates that simply appending existing sentence context to phrasal translations might potentially introduce noise to the training data.

\section{Related Work}


\paragraph{Active Learning for Machine Translation}
Pioneering works on active learning for machine translation focus on selecting sentences that are most useful for training PBMT. This includes sentence selection strategies based on maximizing the percentage of unseen $n$-gram \cite{eck2005low}, $n$-gram frequency, lexical diversity~\cite{haffari-etal-2009-active}, or in-domain coverage~\cite{ananthakrishnan-etal-2010-semi}. 
These sentence selection strategies have been used in active learning algorithms to deal with static data in the batch mode~\cite{ananthakrishnan-etal-2010-semi}, or steaming data in the interactive setting~\cite{gonzalez-rubio-etal-2012-active,peris-casacuberta-2018-active,lam-etal-2019-interactive}. 

For phrase-level annotations, there have been a few works applying phrase-based selection \cite{bloodgood-callison-burch-2010-bucking,miura-etal-2016-selecting} to PBMT. 
While the annotated phrases can be easily integrated by adding them with estimated translation probability to the existing phrase table in PBMT, it it less trivial to integrate these phrase-level annotations in NMT. \citet{arthur-etal-2016-incorporating} integrated the word-level translations to NMT by interpolating the probability of the NMT decoder with the estimated lexical probability. However, this approach requires a modification of the NMT model. Our paper investigates the data-driven approaches that augment the training data by leveraging annotated phrases and existing parallel data.  

\paragraph{Word/Phrase-based Data Augmentation} The other line of research investigates data augmentation methods that leverage word or phrase translations to create synthetic parallel data for training MT models. This includes augmentation methods that replace a word in the existing parallel data with a low-frequency word sampled from the frequency distribution of the vocabulary~\cite{xie2017data} or from the probability of language models in both direction~\cite{fadaee-etal-2017-data,kobayashi-2018-contextual}.
\citet{wang-etal-2018-switchout} proposed an effective method that randomly replaces words in parallel sentences with other random words from the in-domain vocabulary. A more recent work on dictionary-based data augmentation~\cite{peng2020dictionary} proposed to use an existing high-quality in-domain dictionary, and replaced a source word in the existing parallel data by the most similar word in the dictionary according to the cosine similarity metric in the embedding space. In contrast, we select noisy in-domain phrases using different phrase-based selection strategies (\secref{phrase-select}) to ensure the selection quality in an active learning process.


\section{Discussion and Future Work}
\label{sec:conclusion}
In this paper, we investigate ways to incorporating phrasal translations into training NMT for domain adaptation in the active learning setting. We find that phrasal translation is particularly useful in the adaptation scenario where longer sentence context is not necessarily required to translate in-domain words correctly. In contrast, NMT systems can benefit from learning sentence structure with sentence-based selection strategies. The hybrid selection strategies can combine the merits of both sentence-based and phrase-based selection strategies. Nonetheless, there are several future directions. (1) It is worth exploring how different annotation strategies may result in a difference in cost or time. (2) Although several findings could be generalize to other language pairs, testing our methods on morphologically rich languages is our next step. (3) Our current hybrid strategy simply allocate the annotation budget evenly without assuming any prior knowledge on the strategies and the translation performance. Techniques in multi-armed bandit problems~\cite{gittins2011multi} can be used to learn a good allocation strategy. 

\bibliography{reference}
\bibliographystyle{acl_natbib}

\clearpage
\appendix
\section*{Appendix}

\section{Pseudo code}
Algorithm~\ref{alg:active} shows the active learning procedure for machine translation, which consists of two main steps: selection/translation (\cref{sec:selection}) and fine-tuning (\cref{sec:fine-tune}).

\begin{algorithm}[h!]
\caption{Active Learning for Domain Adaptation of Machine Translation}\label{alg:active}
\begin{algorithmic}[1]
\Procedure{ActiveAdaptation}{$\Ucal, \Lcal, B$}
\State \bf{Inputs:} the unlabelled set $\Ucal$, the labelled set $\Lcal$, and a budget $B$.
\State Train a MT model $\theta$ on $\Lcal$.
\State $\Scal, \Pcal \leftarrow$ \textsc{Selection}$(\Ucal, \Lcal, B)$
\State Translate $\Scal$ by $\Lcal_s=\{(x, \Ocal(x))|x\in \Scal\}$
\State Translate $\Pcal$ by $\Lcal_p=\{(p, \Ocal(p))|p\in \Pcal\}$
\State $\Lcal_r \leftarrow$ Obtain parallel data from $\Lcal$ (\secref{fine-tune}) \label{alg:line:retrieve}
\State Fine-tune $\theta$ on $\Lcal_s \cup \Lcal_p \cup \Lcal_r$ \label{alg:line:fine-tune}\\
\Return{$\theta$}
\EndProcedure
\end{algorithmic}
\end{algorithm}

\begin{algorithm}[h!]
\caption{Hybrid Phrase/Sentence Selection 
}\label{alg:selection}
\begin{algorithmic}[1]
\Procedure{Selection}{$\Ucal, \Lcal, B$}
\State \bf{Inputs:} the unlabelled set $\Ucal$, the labelled set $\Lcal$, and a budget $B$.
\State Initialize $\Scal = \{\},~\Pcal = \{\}$
\State Allocate the budget: $B_s, B_p\leftarrow B$ \label{alg:line:allocate}
\While{$\sum_{x\in \Scal}c(x) < B_s$}
    \State $x \leftarrow \argmax_{x\in \Ucal} \phi(x,\cdot)$
    \State $\Ucal = \Ucal \setminus \{x\}$
    \State $\Scal = \Scal \cup \{x\}$
\EndWhile
\State Construct $\Pcal_{\Ucal}, \Pcal_{\Lcal}$ by strategies (\secref{phrase-select})
\While{$\sum_{p \in \Pcal} c(p) < B_p$}
    \State $p \leftarrow \argmax_{p\in \Pcal_{\Ucal}}\occ{p, \Ucal} $
    \State $\Pcal_{\Ucal} = \Pcal_{\Ucal} \setminus \{p\}$
    \State $\Pcal = \Pcal \cup \{p\}$
\EndWhile
\Return{$\Scal, \Pcal$}
\EndProcedure
\end{algorithmic}
\end{algorithm}

\section{Experiments}
\subsection{Experimental Details for Reproducibility}
\label{sec:exp_details}

\paragraph{\textbf{Dataset:}} As pointed out in \citet{aharoni-goldberg-2020-unsupervised}, there is overlap between the training data and the test data in the original split of the two corpora provided by \citet{koehn-knowles-2017-six}, so we follow them in removing the duplicated sentences in the in-domain data, and re-splitting two new test sets in order to prevent the model from memorizing the selected in-domain training data that could potentially be included in the test data. 
Table~\ref{tab:split-de-en} shows the data statistics.
\begin{table}[]
\centering
\resizebox{0.48\textwidth}{!}{%
\begin{tabular}{l|rrrrrr}
\toprule
Data & Domain & Lang & \#Sentences & \#Words & Vocab & Avg Len\\\midrule
\multirow{2}{*}{$\Lcal$} & \multirow{2}{*}{WMT14} & De & \multirow{2}{*}{4.4M} & 108.0M & 1.9M & 24.4\\ 
   &  & En &  & 114.5M & 955.3K & 25.8\\ \midrule
\multirow{2}{*}{$\Ucal$} & Medicine & De & 227.2K & 3.8M & 114.3K & 16.8\\
& IT & De & 190.6K & 2.1M & 114.6K & 11.5 \\ \bottomrule
\end{tabular}%
}
\caption{Data statistics of the out-of-domain labeled data in WMT14 and the in-domain unlabeled data in the medicine and IT domains.} 
\label{tab:split-de-en}
\end{table}

\paragraph{\textbf{Model:}} As our NMT model, we use a 6-layer 512-unit Transformer~\cite{NIPS2017_7181} implemented in \texttt{Fairseq},\footnote{\url{https://github.com/pytorch/fairseq}} and use a subword vocabulary of 5,000 for both languages constructed by Byte Pair Encoding \cite{sennrich-etal-2016-neural}. The model has 45M parameters.

\paragraph{\textbf{Training:}} We train the base model with Adam for 10 epochs with 4K warmup steps and a peak learning rate of 1e-3, and decay the learning rate based on the inverse square root of the number of update steps \cite{NIPS2017_7181}. We save the last checkpoint as our base model, and continue fine-tuning the base model on a mixture of the newly-translated data and the retrieved out-of-domain data for 5 more epochs. 

\paragraph{\textbf{Training/Inference Time:}} We train each model on one NVIDIA RTX 2080Ti GPU for all our experiments. Training the base NMT model takes less than 1 days, and fine-tuning the base NMT model on selected data takes less than 4hours. The decoding of 2000 sentences can be finished within 5 minutes. 

\subsection{How redundant are the selected data?}
To answer this question, we first define ``in-domain words'' as words that only appear in the in-domain test set but do not exist in the out-of-domain data. We report the statistics of the in-domain word types word counts in the selected data with 10K annotated words in Table~\ref{tab:id_word}. We find that phrase selection strategies select more unique in-domain word types and counts than the sentence selection strategies. This indicates that phrase selection strategies leverage the same amount of budget effectively to annotate more diverse in-domain words than sentence selection strategies. 

\begin{table}[]
\centering
\resizebox{0.49\textwidth}{!}{%
\begin{tabular}{lrrr|rrr} \toprule
Methods & IDWT & WT & $\frac{\text{IDWT}}{\text{WT}}$ & IDWC & WC & $\frac{\text{IDWC}}{\text{WC}}$ \\ \midrule
Random Phrase   & 787 & 2206 & 35.68 & 860 & 5003 & 17.19 \\
NGF     & 489 & 1053 & 46.44 & 889 & 5002 & 17.77 \\
NGF-SMP  & 796 & 1492 & 53.35 & 1076 & 5001 & 21.52 \\ \midrule
Random Sentence & 631 & 1984 & 31.80 & 712 & 5023 & 14.17 \\
RTTL    & 592 & 1338 & 44.25 & 961 & 5023 & 19.13 \\
CSSE    & 647 & 2056 & 31.47 & 721 & 5023 & 14.35 \\ \midrule 
NGF-SMP + CSSE   & 667 & 1755 & 38.01 & 859 & 5035 & 17.06 \\ \bottomrule
\end{tabular}%
}
\vspace{-0.2cm}
\caption{Statistics of the unique in-domain word types and word counts in the selected data with 10K annotated words.}
\vspace{-0.4cm}
\label{tab:id_word}
\end{table}

\subsection{Qualitative Analysis}\label{sec:qualitative_sup}

In the first example of Table~\ref{tab:qualitative_sup}, the NMT model adapted by NGF-SMP can predict most words correctly while the NMT model adapted by CSSE generate a random sentence. 

\begin{table*}[]
    \resizebox{0.98\textwidth}{!}{%
    \centering
    \begin{tabular}{l|l@{\hskip -0.5cm}r} \toprule 
         & Output & S-BLEU \\ \midrule
        Source & Schwindel, Parästhesie, Geschmacksstörung &  \\
        Reference & Dizziness, paraesthesiae, taste disorder &  \\
        NGF-SMP & \textblue{Dizziness,} \textred{paraesthesia}, \textblue{taste} \textred{disturbance} & 23.27 \\
        CSSE & \textred{The room was very small and the bathroom was very small.} & 0.00 \\
        NGF-SMP+CSSE &  \textblue{Dizziness,} \textred{paraesthesia}, \textblue{taste} \textred{disturbance} & 23.27 \\ \midrule \midrule
        Source & Über Hospitalisierung oder Todesfälle in Verbindung mit Infektionen wurde berichtet. &  \\
        Reference & Hospitalisation or fatal outcomes associated with infections have been reported. &  \\
        NGF-SMP & \textred{There} have been \textred{reports of} Hospitalisation or \textred{death} associated with infections. & 29.79\\
        CSSE & Hospitals or \textred{deaths} associated with infections have been reported. & 54.63 \\
        NGF-SMP+CSSE & \textred{There} have been \textred{reports of} Hospitalisation or \textred{fatality} associated with infections. & 29.79 \\ \bottomrule 
    \end{tabular}}
    \vspace{-0.2cm}
    \caption{Translations generated by NMT models using different selection strategies. The last column shows the sentence BLEU score of the translations. Translation errors are highlighted in red.}
    \vspace{-0.5cm}
    \label{tab:qualitative_sup}
\end{table*}

\subsection{Do Phrasal Annotations Bias NMT?} 
\begin{figure}
    \centering
    \includegraphics[width=\columnwidth]{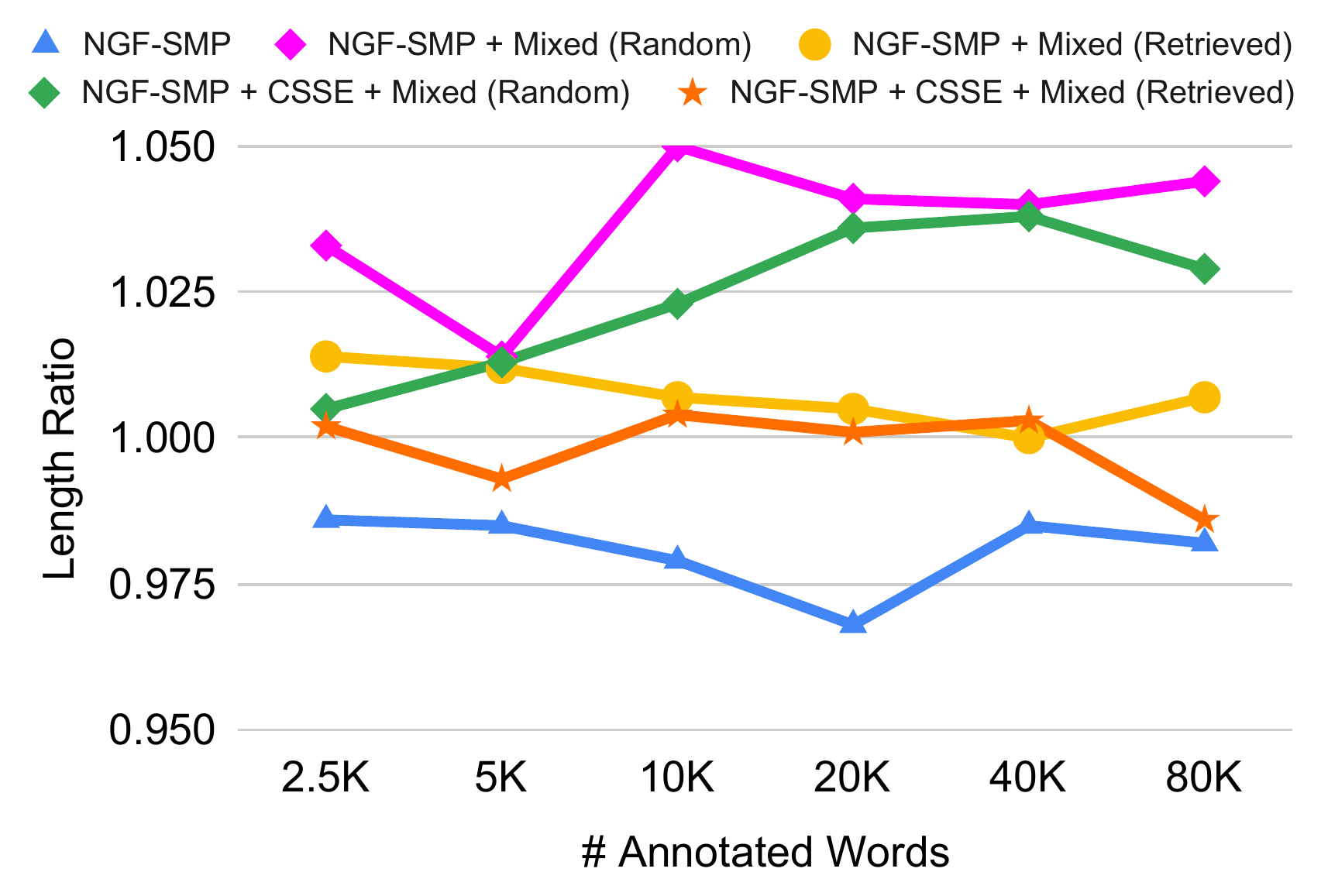}
    \vspace{-0.8cm}
    \caption{Length ratio between the NMT outputs and the reference sentences.} 
    \vspace{-0.4cm}
    \label{fig:length-ratio}
\end{figure}

Since phrasal annotations are short and do not contain complex sentence structure, we hypothesis that NMT systems trained on phrasal annotations would be biased towards generating shorter sentences or sentences in different grammatical order w.r.t. the reference sentence. To understand this question, we analyze the length ratio between the translation outputs and the reference sentences in Figure~\ref{fig:length-ratio}. We find that the NMT model trained only on annotated phrases selected by NGF-SMP generates shorter sentences than reference sentences. In contrast, adding sentences randomly sampled from the labeled corpus $\Lcal$ make the NMT model generate longer sentences than the reference sentences, while retrieving sentences from $\Lcal$ that are similar to the sentences in $\Ucal$ makes the model produces translation outputs with closed lengths as the reference sentences. Qualitatively, we also show the problem of generating sentences with different structures as the reference sentences in the third example in Table~\ref{tab:case_study}. In the third example, the NMT model trained with NGF-SMP produces a translation in an active voice, while the reference sentence uses a passive voice. 

\end{document}